\documentclass{article}

\usepackage[final]{neurips_2023}
\usepackage[utf8]{inputenc} 
\usepackage[T1]{fontenc}    
      
\usepackage{url}            
\usepackage{booktabs}       
\usepackage{amsfonts}       
\usepackage{nicefrac}       
\usepackage{microtype}      
\usepackage{xcolor}         
\usepackage{graphicx}
\usepackage{tabularx}
\usepackage{fontawesome}
\usepackage{xcolor}
\usepackage{caption}
\usepackage{multirow}
\usepackage{textcomp}
\usepackage{hyperref} 

\title{ GeMQuAD \makebox[0pt][l]{\includegraphics[width=0.05\textwidth]{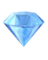}} \hspace{0.6cm}: Generating Multilingual Question Answering Datasets from Large Language Models using Few Shot Learning}

\author{%
  Amani Namboori$^{*}$  \\ 
  Amazon, Alexa International \\
  \texttt{anamburi@amazon.com} \\
  \And
  Shivam Sadashiv Mangale$^{*}$ \\
  Amazon, Alexa International \\
  \texttt{mangsh@amazon.com} \\
  \AND
  Andy Rosenbaum\textsuperscript{\dag} \\
  Amazon, Alexa AI \\
  \texttt{andros@amazon.com} \\
  \And
  Saleh Soltan\textsuperscript{\dag}\\
  Amazon, Alexa AI \\
  \texttt{ssoltan@amazon.com} \\
}
\begin{document}
\maketitle
\begin{abstract}
 The emergence of Large Language Models (LLMs) with capabilities like In-Context Learning (ICL) has ushered in new possibilities for data generation across various domains while minimizing the need for extensive data collection and modeling techniques. Researchers have explored ways to use this generated synthetic data to optimize smaller student models for reduced deployment costs and lower latency in downstream tasks. However, ICL-generated data often suffers from low quality as the task specificity is limited with few examples used in ICL. In this paper, we propose GeMQuAD - a semi-supervised learning approach, extending the WeakDAP framework, applied to a dataset generated through ICL with just one example in the target language using AlexaTM 20B Seq2Seq LLM. Through our approach, we iteratively identify high-quality data to enhance model performance, especially for low-resource multilingual setting in the context of Extractive Question Answering task. Our framework outperforms the machine translation-augmented model by 0.22/1.68 F1/EM (Exact Match) points for Hindi and 0.82/1.37 F1/EM points for Spanish on the MLQA dataset, and it surpasses the performance of model trained on an English-only dataset by 5.05/6.50 F1/EM points   for Hindi and 3.81/3.69 points F1/EM for Spanish on the same dataset. Notably, our approach uses a pre-trained LLM for generation with no fine-tuning (FT), utilizing just a single annotated example in ICL to generate data, providing a cost-effective development process.

\let\thefootnote\relax\footnotetext{* First authors, Equal Contribution}
\let\thefootnote\relax\footnotetext{\dag\ Mentors for this research: Provided strategic vision and offered technical feedback.}
\end{abstract}

\section{Introduction}
While LLMs like ChatGPT \citep{openai-chatgpt}  can answer questions from text, they are computationally expensive and incur huge costs to run at low latency and high throughput. A common alternative is to use a small encoder-only models like BERT\citep{devlin-etal-2019-bert} or XLM-R\citep{xlmrbase2019paper} for extractive QA. However, these smaller models rely on a large quantity of annotated data which is scarce and difficult to obtain, especially for multilingual and domain-specific settings. In the absence of annotated data, synthetic data generation has recently been effective at overcoming scarcity of labeled data for applications including extractive QA \citep{alberti2019synthetic}. 

\begin{figure}[h!]
  \centering
  \includegraphics[width=\linewidth]{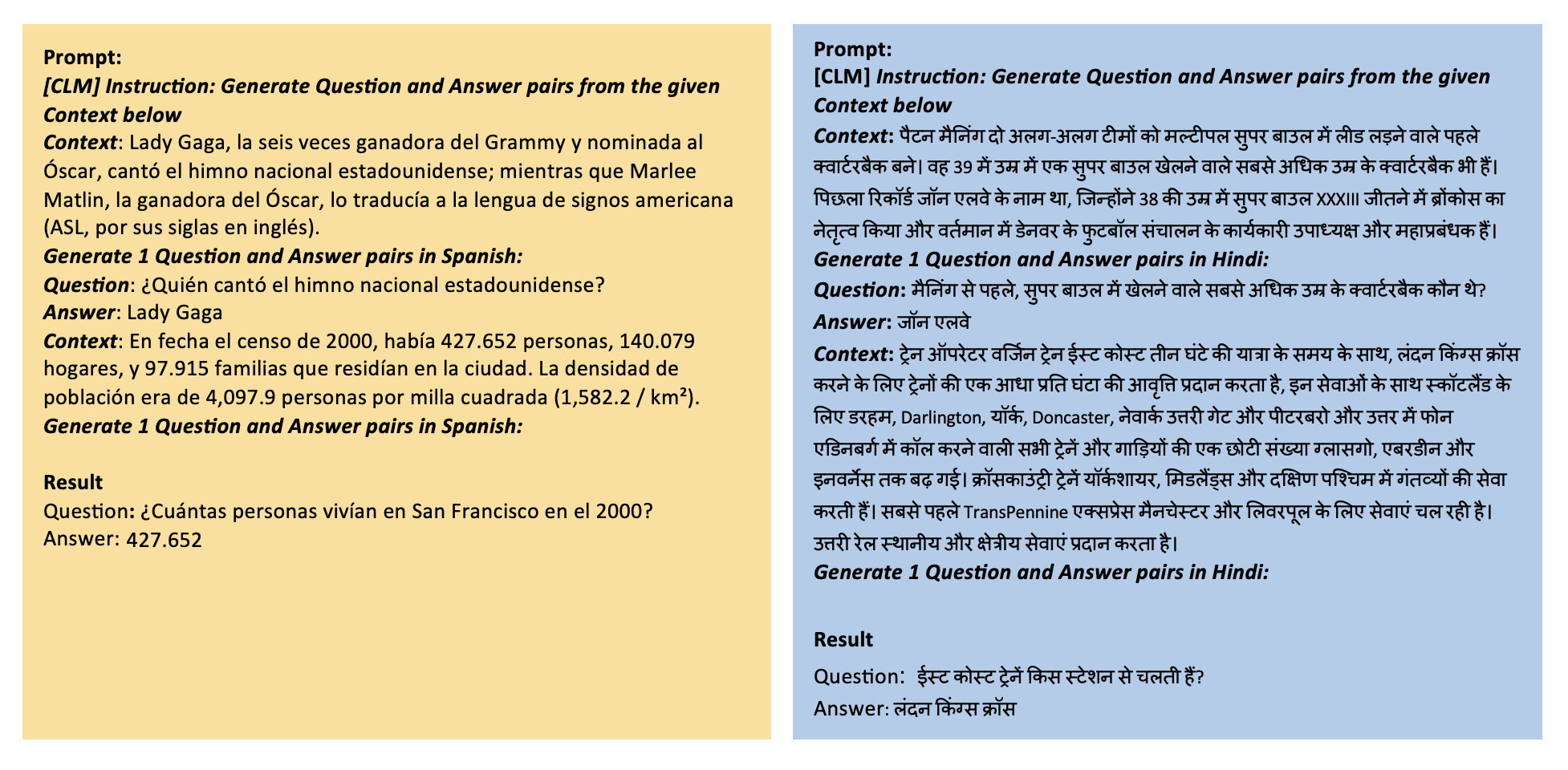}
  \caption{1-Shot example prompts used to generate Spanish and Hindi synthetic data respectively on AlexaTM 20B model using ICL. Prompt instructions are mentioned in English and data is in the target language. For readability, instruction part has been bolded \& italicized. An example synthetic Q\&A pair generated from model presented under Result header. }
  \label{fig:fig1}
\vspace{-3ex}
\end{figure}

Teaching via Data (TvD)\citep{rosenbaum-etal-llms-sdg-2023} involves using a LLM-based teacher model to generate synthetic training data for a specific task, which is used to fine-tune a smaller student model. The challenge is that LLMs, while capable of generating data, lack domain-specific training, and the task representation remains incomplete due to limited examples in ICL. This results in lower data quality compared to annotated data. As demonstrated in \citep{shakeri-etal-2021-towards}, generative models enable task-specific, cross-lingual data generation through fine-tuning on a single language. However, this process is expensive due to the large number of parameters involved in fine-tuning LLMs. 

In this paper, we focus on generating high-quality synthetic data in low-resource scenarios without the need for teacher model fine-tuning. We propose to apply AlexaTM 20B \citep{soltan2022alexatm} to generate synthetic question answering data by prompting the model with a 1 shot in-context example, and we apply a semi supervised approach based on WeakDAP \citep{chen2022weakly} on the student model XLM-R-Base to identify high quality examples from generated data and incrementally improve the performance. While we use AlexaTM 20B as our teacher, this is a generalised framework which can be applied to data generated from any LLM and for any type of task.

\section{Related Work}

Question-Answer pair generation (QAG) has been mostly taken as two independent tasks: given a context to perform Question Generation (QG) and then a Question Answering task (QA). The context used for QG are employed in cross lingual solutions such as generating in English first and then translated using machine translation (MT) into the required language, where \citep{li2023paxqa} did this using alignment, while \citep{fabbri-etal-2020-template} have also looked into template based generation for specific domains. On the other hand \citep{riabi-etal-2021-synthetic} have looked into generating the synthetic data in the target language directly. One difference with such works, is that GeMQuAD doesn't rely on MT  and parallel corpora as we are using multilingual models. QA can be seen as the counterpart of the QG problem, as demonstrated by \citep{kramchaninova-defauw-2022-synthetic} in utilizing context-based candidate answers to create questions. Prior work \citep{shakeri-etal-2021-towards} assesses Question \& Answer (Q\&A)  quality through automatic evaluations, while  \citep{riabi-etal-2021-synthetic} and  \citep{qameleon} have considered downstream QA tasks to quantify their performance.

\textbf{QAMELEON} \citep{qameleon} the most recent related work to ours, also addressed this problem by treating Q\&A generation as a single task. We believe this approach has the potential for better growth compared to treating it as two separate tasks, as it preserves the dependency between questions and answers. Similar to us, QAMELEON proposes using LLMs to generate synthetic Q\&A pairs via ICL with 5 golden samples, which they then use to fine-tune a QA model to evaluate their contributions. The core concept of QAMELEON involves Prompt Tuning (PT) their LLM (PaLM-540B \cite{palm540bpaper}) for the QA task to generate high quality data and evaluate their approach on mt5-XL \cite{mt5xlpaper}. Prompt Tuning (PT) is a parameter efficient fine tuning technique that learns and updates the model parameters specific to input prompt only rather than updating the entire LLM parameters. From a computational perspective, while PT is more effective than FT, it still involves tuning the LLM like PaLM 540B which is of high computational cost. QAMELEON also claims their acheived performance is tied to the base LLM model used i.e., PaLM 540B. However, we believe that using an LLM as large as PaLM is not feasible in low-budget scenarios. Our GeMQuAD approach is a generalised framework which can be applied to any type of synthetic data, to get high quality pairs from the overall generated dataset. We experimented with smaller models with a consideration on low-cost execution,  with no fine-tuning of the generation models, and we are utilizing an iterative approach to develop on our synthetic data as compared to utilizing all the synthetic data generated. To compare our results with QAMELEON, we were unable to replicate the generation strategy because PaLM-540B is a closed model, and the languages in the released QAMELEON dataset do not overlap with our targeted languages.

\section{Methodology}
To improve multilingual performance of the downstream Extractive QA task, our approach consists of 3 major steps. 1) Generate synthetic question answer data using 1-shot ICL on AlexaTM 20B (teacher) model; 2) Apply our semi-supervised learning approach to filter good quality records; 3) fine-tune student model XLM-R-Base with filtered data from step 2. We apply steps 2 \& 3 iteratively until model performance stops improving for a maximum of \textit{k=2} rounds.

\subsection{Synthetic Data Generation using ICL}
Figure \ref{fig:fig1} shows the prompt design used for generating data on AlexaTM 20B using ICL.  \textit{[CLM]} is a special token which the model expects during in-context learning. \textit{Instruction} provides guideline for model to understand that \textit{Question} and \textit{Answer} should be generated from the \textit{Context} which is demonstrated in the 1-shot example. The test context is added after the example. The model learns to generate the question and answer from the test context, in the same format demonstrated in the example. More details on ICL configuration used  is mentioned in the \hyperref[sec:icl_config]{Experimental Setup} section.

\begin{figure}[h!]
\centering
  \includegraphics[width=0.9\linewidth]{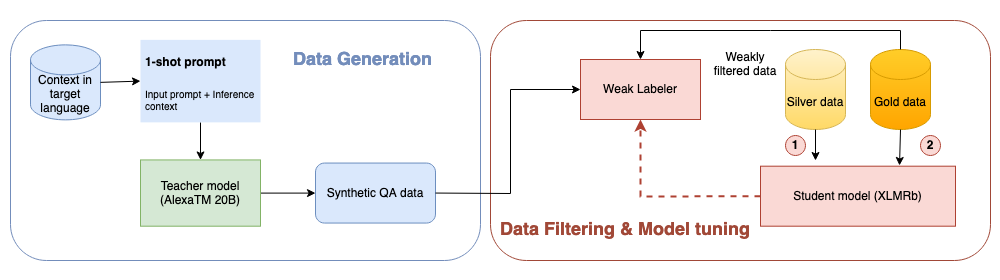}
  \caption{The semi-supervised fine-tuning approach of student using the data generated from teacher. Data generation is a one-time step in which an LLM (AlexaTM) is utilized to create synthetic data using few-shot learning. The data is then passed to the Data Filtering \& Model Tuning stage, where it iteratively filters high-quality records and enhances the labeling model through fine-tuning student (XLM-R-base) until optimal performance is achieved, based on predetermined stopping criteria. }
  \label{fig:fig2}
\vspace{-2ex}
\end{figure}

\subsection{Semi-supervised learning approach}\label{sec:ssl_section}
We are generating data in a low-resource setting, i.e., 1-shot learning. The model generated synthetic data would be of lower quality compared to annotated data. While quality checks can be added to exclude the low quality pairs as part of post processing step, these checks are limited to deterministic and structural validations only. As part of the post-processing, we verify that the answer is a part of the context, and remove any duplicate Q\&A pairs from the dataset. 

To identify good quality functional pairs, we extend the WeakDAP approach with our customization to fine-tuning the student model XLM-R-base for filtering out the \textit{correctly} generated synthetic data. As shown in Figure \ref{fig:fig2}, the workflow begins with the generation of synthetic data in the target language from the teacher model via ICL using the format shown in Figure \ref{fig:fig1}. Once the data generation is complete, a weak labeler evaluates the synthetic data for the Extractive QA task. Given a (context, question) tuple as input, the labeler predicts the answer span within the respective context. 

In our experiments, we initially use our baseline model as the weak labeler, which is trained on  English annotated data corpus (Gold data) for Extractive QA task. As the labeler has not seen any data in target language during training, the Extractive QA performance of the model on target language is limited, and mainly relies on the cross-lingual transfer ability of the base model on which it is fine-tuned. From the synthetic dataset, any questions where our labeler is able to provide the same answer as generated by the teacher model can be considered of high quality and relevant to the context. Consequently, all the Q\&A pairs that the labeler accurately generates based on the synthetic data answer labels are filtered and added to the silver dataset.

\subsection{Student model fine-tuning}

The student model gets fine-tuned first on the data filtered from semi-supervised approach (silver data) and then on the gold data. This is to prioritize the higher quality data in the later part of the FT which we find improves the model performance. Similar analysis is reported by \citep{riabi-etal-2021-synthetic} too in their works.  Student model performance is improved due to the addition of new silver data in target language,  now we replace weak labeler with improved model, and perform \hyperref[sec:ssl_section]{synthetic data evaluation} again. Since the labeler is an improved version compared to its previous iterations, it would be able to identify few more good quality Q\&A pairs from the synthetic dataset which increases the volume of silver data. This process continues iteratively until the model performance doesn't improve for \textit{k} rounds by minimum of \textit{e} threshold (or) the new silver data volume in that iteration is less than \textit{v\%} of the total generated synthetic data. The \textit{k, e, v } values are tunable.

\section{Experimentation}

\subsection{Datasets}
We have used multiple datasets as part of the experimental analysis. In synthetic data generation using ICL, we used the first ten records from the \textbf{XQUAD} \citep{Artetxe_2020} dataset  for Hindi and Spanish labeled examples in our 1-shot ICL prompts. To generate synthetic Q\&A pairs, we used context from the \textbf{XTREME} \citep{hu2020xtreme} 'translate-dev' split. Our student model was trained and validated using the English extractive QA data from \textbf{SQUAD v1.1}  \citep{rajpurkar2016squad}. To assess the student model's performance on multilingual QA tasks, we utilize two evaluation datasets: the \textbf{MLQA} \citep{lewis2019mlqa} dataset and the remaining 1180 records from the XQUAD dataset, excluding the first 10 records. All of these datasets are human-annotated except XTREME, which relies on machine translation. Further details on datasets can be found in \hyperref[section:dataset_details]{Appendix}

\subsection{Experimental Setup}
\subsubsection{ICL Configuration}\label{sec:icl_config}
To perform 1-shot learning on AlexaTM, we have used 10 annotated examples from XQUAD dataset for both Hindi and Spanish data generation, and generated Q\&A pairs for contexts from the XTREME dataset. We randomly pick 1 example to create the input prompt to make sure our generation process is randomised and diversified. On the ICL configuration, we have used sampling approach (\textit{do\_sample= True}), with the \textit{temperature} set to 0.9, and randomly picked \textit{top\_k} \& \textit{top\_p} ranging between 50 to 100 and 0.5 to 0.95 respectively for each example to increase the generalisation in the generated records. The \textit{max\_length} is set to 50 tokens to control generation length.

\subsubsection{Fine-tuning Configuration}

We use the pretrained XLM-R-base \citep{xlmrbase2019paper}  as our student model, fine-tuning it on the QA task with generated synthetic data \& evaluate its performance on both MLQA \& XQUAD datasets. MLQA and XQUAD datasets are both open-domain QA datasets. Therefore, we generated synthetic data only once using the test contexts from XTREME and did not create any synthetic data specific to these datasets. Across all experiments, we consistently use a fixed sample subset of 10,000 records from the SQUAD dataset as gold data, which we refer to as 'SQUAD\textsubscript{en10k}', to maintain consistency in English annotated dataset size with the generated synthetic datasets.

Along with semi-supervised approach (XLMRb\textsubscript{\textit{gemquad}} ), we experimented on the configurations of using the synthetic data in fine-tuning the student model, with XLMRb\textsubscript{\textit{combined}} being the model where we fine-tune it only once on the combined dataset of synthetic data and the SQUAD\textsubscript{\textit{en10k}} subset, while  XLMRb\textsubscript{\textit{sequential}} being the model which is first fine-tuned on complete synthetic data and then fine-tuned on the SQUAD\textsubscript{\textit{en10k}} (similar to our fine-tuning approach but without the semi-supervised filtering). For baseline comparison, we evaluated our approach against two models. XLMRb\textsubscript{\textit{baseline}} was fine-tuned solely on the SQUAD\textsubscript{en10k} subset. We compared our approach with data augmentation using machine translation (MT) \citep{hu2020xtreme}, XLMRb\textsubscript{\textit{MT}} was fine-tuned on a combined dataset that included SQUAD\textsubscript{en10k}, as well as its machine-translated version of Hindi and Spanish data.

XLMRb\textsubscript{\textit{baseline}} is also the initial version of our weak labeler in semi supervised filtering approach. We run loops of iterations, where in each iteration we fine-tune student model on batches of silver data filtered by labeler and SQUAD\textsubscript{en10k}. The improved model in the current iteration evaluates the synthetic data to predict the answers for the given (context, question) pairs, and accordingly classify the synthetic sample as a \textit{correctly} generated example if the answers match or otherwise. The batch of synthetic samples that are classified as \textit{correctly} generated in the current iteration by our student model, are included in silver data for fine-tuning the student model in next iteration. Iterations are stopped if the improvement in model QA task performance for \textit{k=2} rounds by threshold \textit{e<0.005} or the new sample batch volume added to silver data \textit{v<1\%} of total generated synthetic data by ICL. 

For fine-tuning, we have used a linear scheduler with \textit{AdamW}\citep{DBLP:journals/corr/abs-1711-05101} as the optimizer, with \textit{learning\_rate=2e-5} \& \textit{batch\_size=8}. To maintain the consistency of our results across our experiments, we alter the epochs for each experiment so that we have similar number of training update steps so as to make sure the models in each experiment go through similar amount of training in all iterations/experiments. Therefore, any performance changes (taking into account slight fluctuations) relate to the data, validates the use of synthetic data, regardless of its size. We use the best performing epoch according to the validation set in training as our final model at each iteration. 

\section{Results}

\begin{table*}[!ht]
\centering
\begin{tabularx}{\textwidth}{|X|c|c|c|c|}
\hline
\textbf{Model} & \textbf{English} & \textbf{Hindi} & \textbf{Spanish} & \textbf{Average} \\
\hline
XLMRb\textsubscript{\textit{baseline}} & 74.73 / 61.04 & 55.04 / 37.09 & 61.04 / 40.00 & 55.99 / 40.74\\ 
XLMRb\textsubscript{\textit{MT}} & 75.85 /	62.12 & 59.87 /	41.91 & 64.03/ 42.32 & 58.48 /	42.91\\ 
XLMRb\textsubscript{\textit{combined}} & 75.17 / 61.51 & 51.27 / 37.17 & 58.53 / 39.52 & 54.95 / 41.17 \\
XLMRb\textsubscript{\textit{sequential}} & 76.17 / 62.40 & 59.67 / 43.51 & 64.73 / 43.44 & 59.38 / 44.36\\
XLMRb\textsubscript{\textit{gemquad}} & \textbf{76.33 / 62.66} & \textbf{60.09 / 43.59} & \textbf{ 64.85 / 43.69} & \textbf{59.81 / 44.63}\\
\hline
\end{tabularx}
\caption{Performance in QA task for FT with synthetic Hindi \& Spanish data along with SQUAD\textsubscript{en10k} subset. Scores are reported as F1/Exact Match over the MLQA dataset for the 3 languages part of the training and also the average over the languages en, hi, es, de, ar, vi and zh in the MLQA dataset.}
\label{table:table_MLQA}
\end{table*}

Table \ref{table:table_MLQA} \& \ref{table:table_XQUAD} depicts our performance on using synthetic data generated in Hindi and Spanish along with SQUAD\textsubscript{en10k}  on MLQA and XQUAD datasets. We are using the same model to evaluate on both the datasets. On XQUAD, we performed evaluation only to the languages that are common between MLQA \& XQUAD to be consistent on the overall performance averaged across languages. 

\begin{table*}[h]
\centering
\begin{tabularx}{\textwidth}{|X|c|c|c|c|}
\hline
\textbf{Model} & \textbf{English} & \textbf{Hindi} & \textbf{Spanish} & \textbf{Average} \\
\hline
XLMRb\textsubscript{\textit{baseline}} & 78.56 / 66.69	& 61.18 / 45.00  &  71.39 / 51.53  &  67.39 / 51.15 \\ 
XLMRb\textsubscript{\textit{MT}} & \textbf{79.63} / 67.12 & \textbf{67.79} / 50.76 & \textbf{74.22} / 54.15 & 69.39 / 52.96\\ 
XLMRb\textsubscript{\textit{combined}} & 78.95 / 66.44 & 57.85 / 42.88 & 63.05 / 45.17 & 64.67 / 49.41 \\
XLMRb\textsubscript{\textit{sequential}} & 79.36 / 67.54 & 67.18 / 50.76 & 72.79 / 54.24 & 69.97 / 54.36 \\
XLMRb\textsubscript{\textit{gemquad}} &  79.59 / \textbf{67.71 } &	\textbf{67.79 / 52.03 }	& 73.89 / \textbf{55.00} & \textbf{70.49 / 55.10}\\
\hline
\end{tabularx}
\caption{Performance in QA task for FT with synthetic Hindi \& Spanish data along with SQUAD\textsubscript{en10k} subset. Scores are reported as F1/Exact Match over the XQUAD dataset for the 3 languages part of the training and also the average over languages en, hi, es, de, ar, vi and zh in XQUAD dataset.}
\label{table:table_XQUAD}
\vspace{-1ex}
\end{table*}

\begin{figure}[htbp]
    \centering
    \begin{minipage}[h]{0.45\textwidth} 
        \centering
        \begin{tabular}{|c|c|c|}
            \hline
            \textbf{Dataset} & \textbf{\# Q\&A pairs} & \textbf{Type} \\
            \hline
            \rule{0pt}{10pt}
            SQUAD\textsubscript{\textit{en10k}}  & 10000 &  Fine-Tuning\\
            \hline
            \rule{0pt}{10pt}
            Hindi\textsubscript{\textit{MT}}  & 9313 &  Fine-Tuning\\
            \hline
            \rule{0pt}{10pt}
            Spanish\textsubscript{\textit{MT}}  & 9511 &  Fine-Tuning\\
            \hline
            \rule{0pt}{10pt}
            \multirow{2}{*}{Hindi\textsubscript{\textit{syn}}}   & {19558}  &  ICL Generated \&\\
            &  {7395}& Fine-Tuning\\
            \hline
            \rule{0pt}{10pt}
            \multirow{2}{*}{Spanish\textsubscript{\textit{syn}}}  & {15452} & ICL Generated \&\\
            & {7590} & Fine-Tuning\\
            \hline
            \rule{0pt}{10pt}
            MLQA  & 6035 & Evaluation\\
            \hline
            \rule{0pt}{10pt}
            XQUAD  & 1180 & Evaluation\\
            \hline
        \end{tabular}
        \caption{Number of records across datasets and its purpose. Represented average number across languages for MLQA \& XQUAD.}
        \label{table:table_gen_pairs}
    \end{minipage}
    \hfill
    \begin{minipage}{0.45\textwidth} 
        \includegraphics[width=0.95\linewidth]{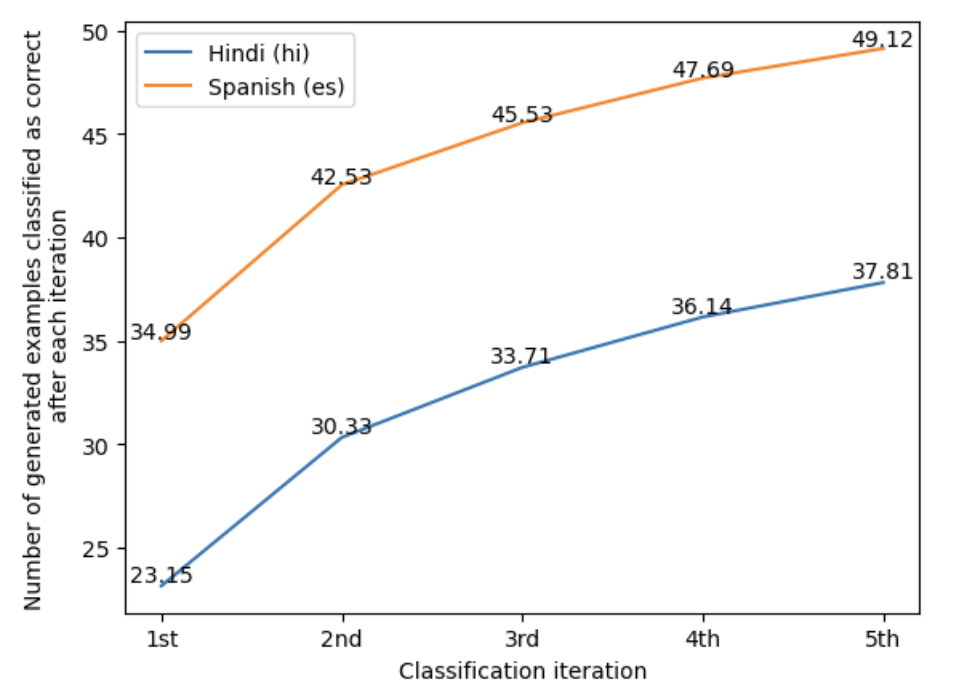}
        \caption{The number of synthetic data samples classified as correct for each iteration }
        \label{fig:fig3}
    \end{minipage}
\end{figure}
\vspace{-1ex}

We generated almost 19.5K Q\&A pairs for Hindi and around 15.5K pairs for Spanish. Table \ref{table:table_gen_pairs} shows these numbers along with volumes of other dataset used in our analysis. Figure \ref{fig:fig3} depicts the number of synthetic data samples classified as correct in each iteration of both Hindi and Spanish experiments. We find that we start with a large share of the samples, of roughly 30\% of the total data, but only gradually add on as we proceed in iterations. This is expected because we only want the best samples, so these are more prone to be selected in the earlier iterations. We are finally able to utilize roughly 45\% of the generated data for our usecase. We achieved our best performance in 3rd iteration. There was no further improvement in 4th and 5th iterations, leading us to stop the iterations with the criterion of k=2. More details on iteration-wise performance is provided in the \hyperref[section:Semi-Supervised-approach-iteration]{Appendix}.

\textbf{Synthetic data helps in performance}  
With our proposed approach(XLMRb\textsubscript{\textit{gemquad}}), we observe similar improvements across both datasets. On MLQA, we find  3.82/3.89 (F1/EM) points of improvement compared to  XLMRb\textsubscript{\textit{baseline}}, with improvement of  5.05/6.50 \& 3.81/3.69 points in Hindi \& Spanish respectively. On XQUAD, we achieved overall improvement of 3.1/3.95 points, with improvement of 6.61/7.03 \& 2.50/3.47 points in Hindi \& Spanish respectively which concludes that having new synthetic examples improves the performance. 

\textbf{Sequentially fine-tuning on separate datasets works better}
Fine-tuning configuration XLMRb\textsubscript{\textit{combined}} yielded mixed results. While English saw a slight performance increase, overall, Hindi and Spanish performance declined. We suspect that by fine-tuning on both synthetic and SQUAD\textsubscript{en10k} together, we may be inadvertently misdirecting the model by giving same importance to all samples. Consequently, we experimented with the XLMRb\textsubscript{\textit{sequential}} configuration, fine-tuning each dataset separately, giving priority to SQUAD\textsubscript{en10k} due to its later fine-tuning which is working better.

\textbf{High-quality, smaller datasets outperform low-quality, large-volume data}
XLMRb\textsubscript{\textit{gemquad}} outperformed XLMRb\textsubscript{\textit{MT}} which involves more data samples than semi-supervised approach. Also, by selecting high quality data pairs from overall dataset worked better than using the entire synthetic dataset generated through ICL (XLMRb\textsubscript{\textit{sequential}})

From our evaluation above, overall GeMQuAD outperformed XLMRb\textsubscript{\textit{combined}} by an improvement of 5.3/4.6 F1/EM (Average performance across all languages in both MLQA \& XQUAD datasets) and XLMRb\textsubscript{\textit{sequential}} by 0.5/0.5 F1/EM. On the baselines, our approach works better than machine translation XLMRb\textsubscript{\textit{MT}}  by 1.2/1.9 F1/EM,  while using only English annotated data  XLMRb\textsubscript{\textit{baseline}} by  3.5/3.9 F1/EM. In short, \texttt{GeMQuAD > sequential (0.5/0.5) > MT (1.2/1.9) > baseline (3.5/3.9) > combined (5.3/4.6)}.

We observed good improvements in languages that are not included in the student model's fine-tuning data, such as German, Arabic, Vietnamese, and Chinese. Detailed information on performance of these languages provided in the \hyperref[section:model_performance_on_all_languages]{Appendix}. We attribute this enhanced cross-lingual ability to our selection of higher-quality samples from our generated synthetic data.  

\section{Conclusion \& Next steps}
In this paper, we introduced GeMQuAD, a cost-effective approach for generating synthetic data in Hindi and Spanish using ICL on AlexaTM. We demonstrated the effectiveness of a semi-supervised method for selecting high-quality synthetic Q\&A pairs, resulting in an improvement of 0.22/1.68 F1/EM for Hindi and 0.82/1.37 F1/EM for Spanish compared to the model trained on English data, augmented with machine translation. Our approach outperformed the model trained solely on an English-only dataset by 5.05/6.50 F1/EM for Hindi and 3.81/3.69 F1/EM for Spanish on the MLQA dataset, all achieved without fine-tuning the LLM. Our next steps involve extending this analysis to other low-resource languages, performing domain-specific QA, and expanding the functionality to include abstractive QA generation.

\bibliography{main}

\begin{thebibliography}{20}
\expandafter\ifx\csname natexlab\endcsname\relax\def\natexlab#1{#1}\fi

\bibitem[{Agrawal et~al.(2023)Agrawal, Alberti, Huot, Maynez, Ma, Ruder,
  Ganchev, Das, and Lapata}]{qameleon}
Priyanka Agrawal, Chris Alberti, Fantine Huot, Joshua Maynez, Ji~Ma, Sebastian
  Ruder, Kuzman Ganchev, Dipanjan Das, and Mirella Lapata. 2023.
\newblock \href {http://arxiv.org/abs/2211.08264} {Qameleon: Multilingual qa
  with only 5 examples}.

\bibitem[{Alberti et~al.(2019)Alberti, Andor, Pitler, Devlin, and
  Collins}]{alberti2019synthetic}
Chris Alberti, Daniel Andor, Emily Pitler, Jacob Devlin, and Michael Collins.
  2019.
\newblock \href {http://arxiv.org/abs/1906.05416} {Synthetic qa corpora
  generation with roundtrip consistency}.

\bibitem[{Artetxe et~al.(2020)Artetxe, Ruder, and Yogatama}]{Artetxe_2020}
Mikel Artetxe, Sebastian Ruder, and Dani Yogatama. 2020.
\newblock \href {https://doi.org/10.18653/v1/2020.acl-main.421} {On the
  cross-lingual transferability of monolingual representations}.
\newblock In \emph{Proceedings of the 58th Annual Meeting of the Association
  for Computational Linguistics}. Association for Computational Linguistics.

\bibitem[{Chen et~al.(2022)Chen, Papangelis, Tao, Rosenbaum, Kim, Liu, Yu, and
  Hakkani-Tur}]{chen2022weakly}
Maximillian Chen, Alexandros Papangelis, Chenyang Tao, Andy Rosenbaum, Seokhwan
  Kim, Yang Liu, Zhou Yu, and Dilek Hakkani-Tur. 2022.
\newblock \href {http://arxiv.org/abs/2210.14169} {Weakly supervised data
  augmentation through prompting for dialogue understanding}.

\bibitem[{Chowdhery et~al.(2022)Chowdhery, Narang, Devlin, Bosma, Mishra,
  Roberts, Barham, Chung, Sutton, Gehrmann, Schuh, Shi, Tsvyashchenko, Maynez,
  Rao, Barnes, Tay, Shazeer, Prabhakaran, Reif, Du, Hutchinson, Pope, Bradbury,
  Austin, Isard, Gur-Ari, Yin, Duke, Levskaya, Ghemawat, Dev, Michalewski,
  Garcia, Misra, Robinson, Fedus, Zhou, Ippolito, Luan, Lim, Zoph, Spiridonov,
  Sepassi, Dohan, Agrawal, Omernick, Dai, Pillai, Pellat, Lewkowycz, Moreira,
  Child, Polozov, Lee, Zhou, Wang, Saeta, Diaz, Firat, Catasta, Wei,
  Meier-Hellstern, Eck, Dean, Petrov, and Fiedel}]{palm540bpaper}
Aakanksha Chowdhery, Sharan Narang, Jacob Devlin, Maarten Bosma, Gaurav Mishra,
  Adam Roberts, Paul Barham, Hyung~Won Chung, Charles Sutton, Sebastian
  Gehrmann, Parker Schuh, Kensen Shi, Sasha Tsvyashchenko, Joshua Maynez,
  Abhishek Rao, Parker Barnes, Yi~Tay, Noam Shazeer, Vinodkumar Prabhakaran,
  Emily Reif, Nan Du, Ben Hutchinson, Reiner Pope, James Bradbury, Jacob
  Austin, Michael Isard, Guy Gur-Ari, Pengcheng Yin, Toju Duke, Anselm
  Levskaya, Sanjay Ghemawat, Sunipa Dev, Henryk Michalewski, Xavier Garcia,
  Vedant Misra, Kevin Robinson, Liam Fedus, Denny Zhou, Daphne Ippolito, David
  Luan, Hyeontaek Lim, Barret Zoph, Alexander Spiridonov, Ryan Sepassi, David
  Dohan, Shivani Agrawal, Mark Omernick, Andrew~M. Dai,
  Thanumalayan~Sankaranarayana Pillai, Marie Pellat, Aitor Lewkowycz, Erica
  Moreira, Rewon Child, Oleksandr Polozov, Katherine Lee, Zongwei Zhou, Xuezhi
  Wang, Brennan Saeta, Mark Diaz, Orhan Firat, Michele Catasta, Jason Wei,
  Kathy Meier-Hellstern, Douglas Eck, Jeff Dean, Slav Petrov, and Noah Fiedel.
  2022.
\newblock \href {http://arxiv.org/abs/2204.02311} {Palm: Scaling language
  modeling with pathways}.

\bibitem[{Conneau et~al.(2019)Conneau, Khandelwal, Goyal, Chaudhary, Wenzek,
  Guzm{\'{a}}n, Grave, Ott, Zettlemoyer, and Stoyanov}]{xlmrbase2019paper}
Alexis Conneau, Kartikay Khandelwal, Naman Goyal, Vishrav Chaudhary, Guillaume
  Wenzek, Francisco Guzm{\'{a}}n, Edouard Grave, Myle Ott, Luke Zettlemoyer,
  and Veselin Stoyanov. 2019.
\newblock \href {http://arxiv.org/abs/1911.02116} {Unsupervised cross-lingual
  representation learning at scale}.
\newblock \emph{CoRR}, abs/1911.02116.

\bibitem[{Devlin et~al.(2019)Devlin, Chang, Lee, and
  Toutanova}]{devlin-etal-2019-bert}
Jacob Devlin, Ming-Wei Chang, Kenton Lee, and Kristina Toutanova. 2019.
\newblock \href {https://doi.org/10.18653/v1/N19-1423} {{BERT}: Pre-training of
  deep bidirectional transformers for language understanding}.
\newblock In \emph{Proceedings of the 2019 Conference of the North {A}merican
  Chapter of the Association for Computational Linguistics: Human Language
  Technologies, Volume 1 (Long and Short Papers)}, pages 4171--4186,
  Minneapolis, Minnesota. Association for Computational Linguistics.

\bibitem[{Fabbri et~al.(2020)Fabbri, Ng, Wang, Nallapati, and
  Xiang}]{fabbri-etal-2020-template}
Alexander Fabbri, Patrick Ng, Zhiguo Wang, Ramesh Nallapati, and Bing Xiang.
  2020.
\newblock \href {https://doi.org/10.18653/v1/2020.acl-main.413} {Template-based
  question generation from retrieved sentences for improved unsupervised
  question answering}.
\newblock In \emph{Proceedings of the 58th Annual Meeting of the Association
  for Computational Linguistics}, pages 4508--4513, Online. Association for
  Computational Linguistics.

\bibitem[{Hu et~al.(2020)Hu, Ruder, Siddhant, Neubig, Firat, and
  Johnson}]{hu2020xtreme}
Junjie Hu, Sebastian Ruder, Aditya Siddhant, Graham Neubig, Orhan Firat, and
  Melvin Johnson. 2020.
\newblock \href {https://proceedings.mlr.press/v119/hu20b.html} {{XTREME}: A
  massively multilingual multi-task benchmark for evaluating cross-lingual
  generalisation}.
\newblock In \emph{Proceedings of the 37th International Conference on Machine
  Learning}, volume 119 of \emph{Proceedings of Machine Learning Research},
  pages 4411--4421. PMLR.

\bibitem[{Kramchaninova and Defauw(2022)}]{kramchaninova-defauw-2022-synthetic}
Alina Kramchaninova and Arne Defauw. 2022.
\newblock \href {https://aclanthology.org/2022.eamt-1.18} {Synthetic data
  generation for multilingual domain-adaptable question answering systems}.
\newblock In \emph{Proceedings of the 23rd Annual Conference of the European
  Association for Machine Translation}, pages 151--160, Ghent, Belgium.
  European Association for Machine Translation.

\bibitem[{Lewis et~al.(2019)Lewis, O\u{g}uz, Rinott, Riedel, and
  Schwenk}]{lewis2019mlqa}
Patrick Lewis, Barlas O\u{g}uz, Ruty Rinott, Sebastian Riedel, and Holger
  Schwenk. 2019.
\newblock Mlqa: Evaluating cross-lingual extractive question answering.
\newblock \emph{arXiv preprint arXiv:1910.07475}.

\bibitem[{Li and Callison-Burch(2023)}]{li2023paxqa}
Bryan Li and Chris Callison-Burch. 2023.
\newblock \href {http://arxiv.org/abs/2304.12206} {Paxqa: Generating
  cross-lingual question answering examples at training scale}.

\bibitem[{Loshchilov and Hutter(2017)}]{DBLP:journals/corr/abs-1711-05101}
Ilya Loshchilov and Frank Hutter. 2017.
\newblock \href {http://arxiv.org/abs/1711.05101} {Fixing weight decay
  regularization in adam}.
\newblock \emph{CoRR}, abs/1711.05101.

\bibitem[{{OpenAI}(2023)}]{openai-chatgpt}
{OpenAI}. 2023.
\newblock {GPT-3.5 (ChatGPT)}.
\newblock \url{https://openai.com}.

\bibitem[{Rajpurkar et~al.(2016)Rajpurkar, Zhang, Lopyrev, and
  Liang}]{rajpurkar2016squad}
Pranav Rajpurkar, Jian Zhang, Konstantin Lopyrev, and Percy Liang. 2016.
\newblock \href {http://arxiv.org/abs/1606.05250} {Squad: 100,000+ questions
  for machine comprehension of text}.

\bibitem[{Riabi et~al.(2021)Riabi, Scialom, Keraron, Sagot, Seddah, and
  Staiano}]{riabi-etal-2021-synthetic}
Arij Riabi, Thomas Scialom, Rachel Keraron, Beno{\^\i}t Sagot, Djam{\'e}
  Seddah, and Jacopo Staiano. 2021.
\newblock \href {https://doi.org/10.18653/v1/2021.emnlp-main.562} {Synthetic
  data augmentation for zero-shot cross-lingual question answering}.
\newblock In \emph{Proceedings of the 2021 Conference on Empirical Methods in
  Natural Language Processing}, pages 7016--7030, Online and Punta Cana,
  Dominican Republic. Association for Computational Linguistics.

\bibitem[{Rosenbaum et~al.(2023)Rosenbaum, Soltan, and
  Hamza}]{rosenbaum-etal-llms-sdg-2023}
Andy Rosenbaum, Saleh Soltan, and Wael Hamza. 2023.
\newblock \href
  {https://www.amazon.science/blog/using-large-language-models-llms-to-synthesize-training-data}
  {Using large language models (llms) to synthesize training data}.
\newblock \emph{Amazon Science}.

\bibitem[{Shakeri et~al.(2021)Shakeri, Constant, Kale, and
  Xue}]{shakeri-etal-2021-towards}
Siamak Shakeri, Noah Constant, Mihir Kale, and Linting Xue. 2021.
\newblock \href {https://aclanthology.org/2021.inlg-1.4} {Towards zero-shot
  multilingual synthetic question and answer generation for cross-lingual
  reading comprehension}.
\newblock In \emph{Proceedings of the 14th International Conference on Natural
  Language Generation}, pages 35--45, Aberdeen, Scotland, UK. Association for
  Computational Linguistics.

\bibitem[{Soltan et~al.(2022)Soltan, Ananthakrishnan, FitzGerald, Gupta, Hamza,
  Khan, Peris, Rawls, Rosenbaum, Rumshisky, Prakash, Sridhar, Triefenbach,
  Verma, Tur, and Natarajan}]{soltan2022alexatm}
Saleh Soltan, Shankar Ananthakrishnan, Jack FitzGerald, Rahul Gupta, Wael
  Hamza, Haidar Khan, Charith Peris, Stephen Rawls, Andy Rosenbaum, Anna
  Rumshisky, Chandana~Satya Prakash, Mukund Sridhar, Fabian Triefenbach, Apurv
  Verma, Gokhan Tur, and Prem Natarajan. 2022.
\newblock \href {http://arxiv.org/abs/2208.01448} {Alexatm 20b: Few-shot
  learning using a large-scale multilingual seq2seq model}.

\bibitem[{Xue et~al.(2021)Xue, Constant, Roberts, Kale, Al-Rfou, Siddhant,
  Barua, and Raffel}]{mt5xlpaper}
Linting Xue, Noah Constant, Adam Roberts, Mihir Kale, Rami Al-Rfou, Aditya
  Siddhant, Aditya Barua, and Colin Raffel. 2021.
\newblock \href {http://arxiv.org/abs/2010.11934} {mt5: A massively
  multilingual pre-trained text-to-text transformer}.

\end{thebibliography}
\bibliographystyle{acl_natbib}
\pagebreak

\appendix
\section*{Appendix}
\subsection*{A Model performance on other languages}
\label{section:model_performance_on_all_languages}
The tables \ref{table:table_MLQA_other} and \ref{table:table_XQUAD_other}  presents model performance and comparison for languages in MLQA and XQUAD datasets, excluding English, Hindi, and Spanish. While not part of fine-tuning, we observed notable improvement in these languages. This suggests that as the model's overall performance improves with additional data in target languages, its cross-lingual transfer capability also increases, leading to enhanced performance in these languages. We limited the evaluation to 7 languages common with MLQA to maintain consistency in cross-language comparisons, although XQUAD supports 12 languages.

\begin{table}[h]
\centering
\begin{tabularx}{\textwidth}{|X|c|c|c|c|}
\hline
\textbf{Model} & \textbf{German} & \textbf{Arabic} & \textbf{Vietnamese} & \textbf{Chinese} \\
\hline
XLMRb\textsubscript{\textit{baseline}} & 56.23 / 41.18 & 49.56 / 31.10 & 61.13 / 41.18 & 34.21 / 33.62\\ 
XLMRb\textsubscript{\textit{MT}} & 58.00 / 42.02 & 50.57 / 32.15 & 63.22 / 42.58 & 37.83 / 37.26\\
XLMRb\textsubscript{\textit{combined}} & 56.06 / 41.89 & 46.94 / 31.08 & 60.69 / 41.64 & 35.96 / 35.39 \\
XLMRb\textsubscript{\textit{sequential}} & 59.26 / 44.12 & 52.53 / 34.43 & 64.60 / 44.59 & 38.67 / 38.06\\
XLMRb\textsubscript{\textit{gemquad}} & 59.32 / 43.57 & 52.56 / 34.41 & 66.13 / 45.77 & 39.37 / 38.74\\
\hline
\end{tabularx}
\linebreak
\linebreak
\caption{Performance in QA task for FT with synthetic Hindi \& Spanish data along with SQUAD\textsubscript{en10k} subset. Scores are reported as F1/Exact Match over the MLQA dataset for the  languages that are not part of the FT, such as German, Arabic, Vietnamese, and Chinese.}
\label{table:table_MLQA_other}
\end{table}

\begin{table}[h]
\centering
\begin{tabularx}{\textwidth}{|X|c|c|c|c|}
\hline
\textbf{Model} & \textbf{German} & \textbf{Arabic} & \textbf{Vietnamese} & \textbf{Chinese} \\
\hline
XLMRb\textsubscript{\textit{baseline}} & 69.09 / 52.29 & 62.49 / 45.76 & 69.10 / 49.07 & 59.95 / 47.71\\ 
XLMRb\textsubscript{\textit{MT}} & 69.33 / 51.95 & 62.41 / 44.66 & 71.18 / 50.67 & 61.19 / 51.44\\
XLMRb\textsubscript{\textit{combined}} & 67.50 / 51.95 & 57.42 / 40.51 & 66.74 / 46.95 & 61.16 / 51.95 \\
XLMRb\textsubscript{\textit{sequential}} & 71.64 / 55.34 & 64.08 / 47.20 & 70.76 / 50.93 & 63.96 / 54.49\\
XLMRb\textsubscript{\textit{gemquad}} & 70.92 / 55.17 & 64.96 / 48.22 & 72.03 / 51.86 & 64.23 / 55.68\\
\hline
\end{tabularx}
\linebreak
\linebreak
\caption{Performance in QA task for FT with synthetic Hindi \& Spanish data along with SQUAD\textsubscript{en10k} subset. Scores are reported as F1/Exact Match over the XQUAD dataset for the  languages that are not part of the FT, such as German, Arabic, Vietnamese, and Chinese.}
\label{table:table_XQUAD_other}
\end{table}

\subsection*{B Semi-Supervised approach iteration details}\label{section:Semi-Supervised-approach-iteration}
This section provides detailed insights into the incremental performance \& data included fine-tuning across iterations of semi-supervised approach proposed in this paper.

For the experiment with fine-tuning student model XLM-R-Base with synthetic data generated from AlexaTM 20B in Hindi \& Spanish, our semi-supervised iterative approach concluded at 5 iterations, with 3rd iteration being the best in performance as the model performance didn't improve on 4th and 5th iteration (i.e., k=2).

\begin{table}[h]
\centering
\begin{tabularx}{\textwidth}{|c|X|X|c|c|c|}
\hline
\textbf{Iteration} & \textbf{Hindi Data} & \textbf{Spanish Data} & \textbf{SQUAD\textsubscript{\textit{validation}}} & \textbf{MLQA} & \textbf{XQUAD} \\
\hline
Round 1 & 4528 & 5407 & 84.23 / 76.27 & 58.24 / 43.53 & 69.00 / 53.90\\ 
Round 2  & 5931	& 6572 & 84.36 / 76.73 & 58.90 / 43.64 & 69.56 / 55.17 \\
Round 3 & 6593	& 7035 & 85.07 / 77.26 & 59.81 / 44.63 & 70.49 / 55.10\\
Round 4 & 7069	& 7369 & 84.96 / 76.87 & 59.18 / 43.85 & 69.76 / 54.20\\
Round 5 & 7395	& 7590 & 84.72 / 76.45 & 58.89 / 43.61 & 69.71 / 54.08\\
\hline
\end{tabularx}
\linebreak
\linebreak
\caption{Distribution of data samples included in semi-supervised iterative approach. Each round identifies the additional good quality Q\&A pairs from synthetic data and adds to the silver sets of Hindi \& Spanish datasets for student model training. Gold data (English annotated) is constant across all iterations which is 10k SQuAD\textsubscript{en}. Right most 2 columns represent average performance across all 7 languages on MLQA \& XQUAD datasets }
\label{table:table_semisupervised}
\end{table}

Table \ref{table:table_semisupervised} represents the synthetic data distribution included in semi-supervised iterations. Overall number of samples included in each round includes synthetic Hindi data, synthetic Spanish data and  SQUAD\textsubscript{en10k}. For example for 1st round, overall data included is 4528+5407+10000 = 19935. During training, model performance continuously improved till round 3 on SQUAD\textsubscript{\textit{validation}} (validation dataset used in student model fine-tuning), post which there is a slight decline for 2 rounds. As per the stopping criteria, we stop the iterative process, and pick round 3 as the best performing model from the semi-supervised approach. Along with model performance on validation set, we represented model performance on MLQA and XQUAD evaluation datasets as well for reference. Same performance pattern is visible on both evaluation datasets.

\begin{figure}[htbp]
    \centering
    \begin{minipage}{0.45\textwidth} 
        \includegraphics[width=\linewidth]{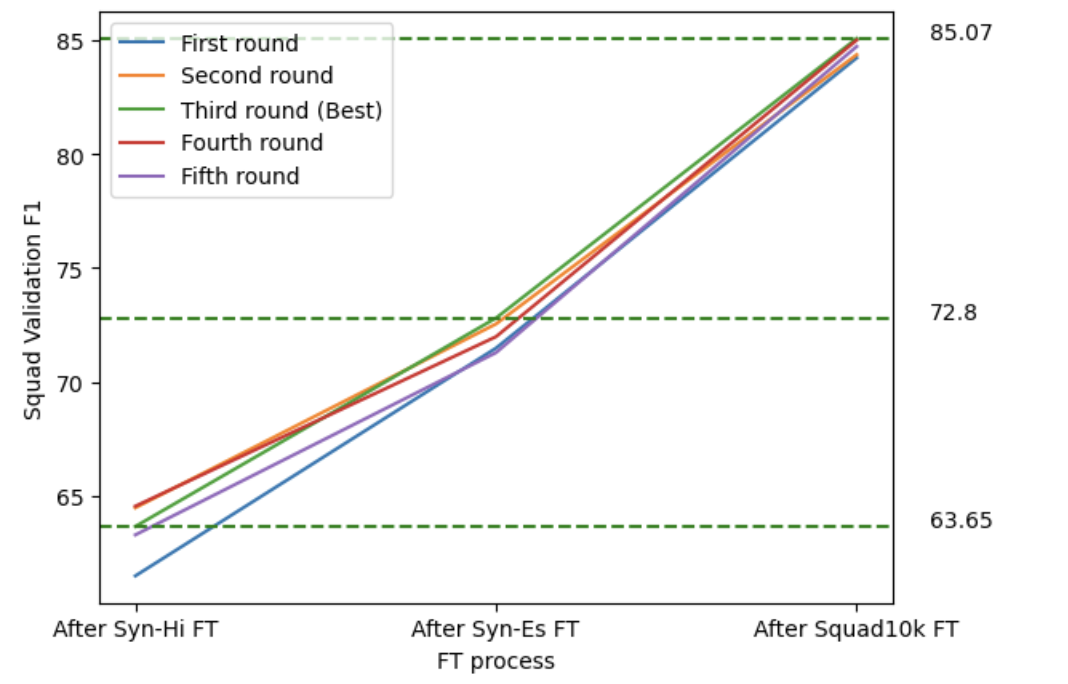}
        \caption{The F1 for Squad Validation split after each FT step. The horizontal green dashed lines signify the performance of the Best (Third) round in the at each step.}
        \label{fig:rounds_f1}
    \end{minipage}
    \hfill
    \begin{minipage}{0.45\textwidth} 
        \includegraphics[width=\linewidth]{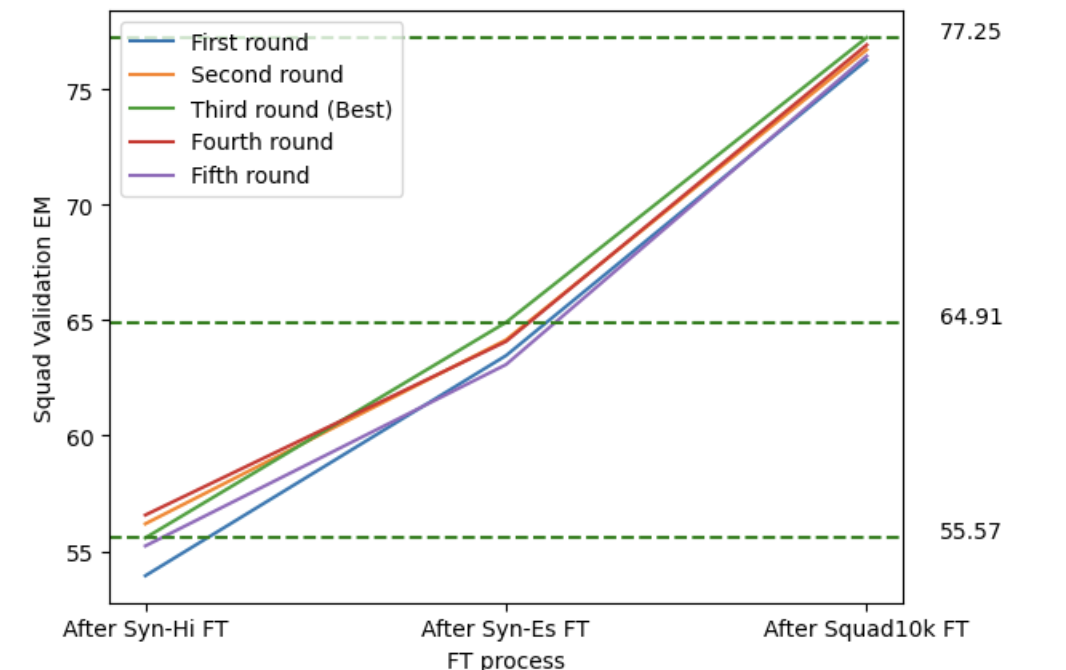}
        \caption{The EM for Squad Validation split after each FT step. The horizontal green dashed lines signify the performance of the Best (Third) round in the at each step.}
        \label{fig:rounds_em}
    \end{minipage}
\end{figure}

\label{section:perf_imp_iteration_val}
\textbf{Performance improvement in iterations} Figures \ref{fig:rounds_f1} and \ref{fig:rounds_em} show the validation performance (SQUAD\textsubscript{\textit{validation}}) of the model during fine-tuning across the iterations. As explained in the paper, our fine-tuning framework is a multi step process. The model gets trained on Hindi synthetic data first, Spanish synthetic data second and English annotated data as the last step to keep high quality data in the later phases. The above 2 plots show the validation performance that the model achieved at each step. An interesting pattern here is that as the overall model performance increases from round 1 to 3, it is also reflected in the individual steps. For example, the model performance increased in Hindi from round 1 (53.98) to round 2 (56.18) for Exact Match. A similar pattern can be observed for Spanish and English steps as well for both EM and F1 plots. With round 3 being highest for Spanish, the performance in round 4 and 5 is less than 3. The same pattern is observed on our test sets MLQA \& XQUAD performance too. This enables the understanding that the performance at each step improves the performance of next step and subsequently overall performance. So having good samples as part of each step is essential to get the optimum performance.

\section*{C Datasets}\label{section:dataset_details}
\textbf{SQUAD v1.1} \citep{rajpurkar2016squad} is an extractive QA dataset in English language that has been created from more than 107k context-question-answer triplets across 536 articles.These articles have been taken from Wikipedia and the dataset is annotated using MechanicalTurk\footnote{https://www.mturk.com/}.
This is the most common dataset used by many studies related to QA tasks. SQUAD dataset has 2 subsets, in which the train subset is used for training the baseline model on labelled English data. The validation subset is used as the evaluation dataset for XLM-R-Base (student model) training process. 

\textbf{XQUAD} \citep{Artetxe_2020} is an extractive QA dataset which is a multilingual version of SQUAD. This dataset has only 1 subset, a human annotated validation set of 1190 examples in 12 supported languages. We have used the first 10 records of Hindi and Spanish XQUAD subset as annotated examples in our 1-shot ICL prompts. Remaining 1180 records are used to compare the performance of the downstream task of extractive QA across languages on student model. Even though we use only 1 example at a time to generate the data, we have randomised the input prompt from these 10 examples to add diversity.

\textbf{XTREME} \citep{hu2020xtreme} is a multilingual extractive QA dataset created by translating the SQUADv1.1  using machine translation.  To generate synthetic Q\&A pairs for hi (Hindi) and es (Spanish) languages, we use context information from translate-dev split of Xquad-Xtreme\footnote{https://huggingface.co/datasets/juletxara/xquad\_xtreme/viewer/hi/translate\_dev} version dataset as our test contexts, i.e., the contexts for which the Q\&A pairs are generated. We have not used the labelled Q\&A pairs available in this dataset for any of our training purposes.

\textbf{MLQA} \citep{lewis2019mlqa} is an extractive multilingual QA benchmark dataset that has been manually created. From language specific subsets, we use the test split to understand and report our performance on downstream task of extractive QA across languages on student model.

\end{document}